\documentclass{article}

\usepackage[numbers]{natbib}

\usepackage[final]{neurips_2024}

\usepackage[utf8]{inputenc} 
\usepackage[T1]{fontenc}    
\usepackage{hyperref}       
\usepackage{url}            
\usepackage{booktabs}       
\usepackage{amsfonts}       
\usepackage{nicefrac}       
\usepackage{microtype}      
\usepackage{xcolor}         
\usepackage{tabularx}
\usepackage{array}  
\usepackage{graphicx}
\usepackage{comment}
\usepackage{float}
\usepackage{listings}
\usepackage{caption}
\usepackage{xcolor}

\definecolor{gray}{rgb}{0.5, 0.5, 0.5}
\definecolor{lightgray}{rgb}{0.95, 0.95, 0.95}

\newcolumntype{C}{>{\centering\arraybackslash}X}
\newcolumntype{Y}{>{\centering\arraybackslash}X}

\title{Synthetic SQL Column Descriptions and Their Impact on Text-to-SQL Performance}

\author{%
  Niklas Wretblad*, Oskar Holmström, Erik Larsson, Axel Wiksäter, Hjalmar Öhman,\\
  \textbf{Oscar Söderlund, Ture Pontén, Martin Forsberg, Martin Sörme, Fredrik Heintz} \\
  \\
  Department of Information and Computer Science\\
  Linköping University, Linköping, Sweden \\
  \texttt{*niklas.wretblad@liu.se} \\
}

\begin{document}

\maketitle
\raggedbottom

\begin{abstract}

Relational databases often suffer from uninformative descriptors of table contents, such as ambiguous columns and hard-to-interpret values, impacting both human users and text-to-SQL models. In this paper, we explore the use of large language models (LLMs) to automatically generate detailed natural language descriptions for SQL database columns, aiming to improve text-to-SQL performance and automate metadata creation. We create a dataset of gold column descriptions based on the BIRD-Bench benchmark, manually refining its column descriptions and creating a taxonomy for categorizing column difficulty. We then evaluate several different LLMs in generating column descriptions across the columns and different difficulties in the dataset, finding that models unsurprisingly struggle with columns that exhibit inherent ambiguity, highlighting the need for manual expert input. We also find that incorporating such generated column descriptions consistently enhances text-to-SQL model performance, particularly for larger models like GPT-4o, Qwen2 72B and Mixtral 22Bx8.  Notably, Qwen2-generated descriptions, containing by annotators deemed superfluous information, outperform manually curated gold descriptions, suggesting that models benefit from more detailed metadata than humans expect. Future work will investigate the specific features of these high-performing descriptions and explore other types of metadata, such as numerical reasoning and synonyms, to further improve text-to-SQL systems. The dataset, annotations and code will all be made available. 
\end{abstract}

\section{Introduction}

\begin{figure}[t] 
  \centering
  \includegraphics[width=\linewidth]{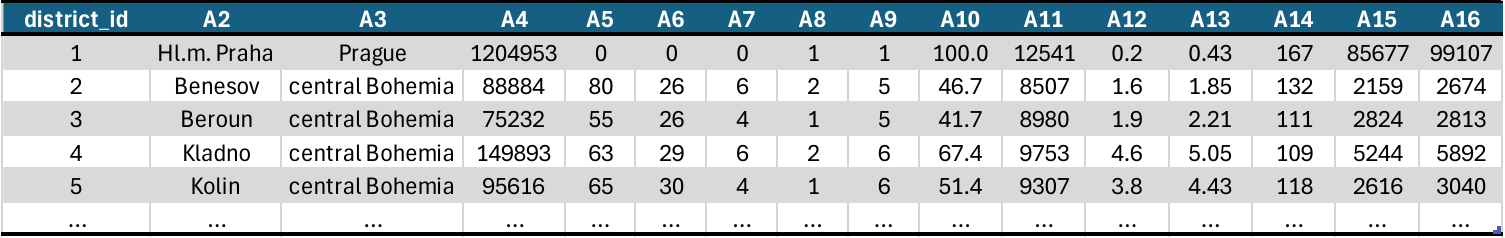}
  \caption{Example rows from the \texttt{district} table in the BIRD-Bench benchmark dataset. This table illustrates a typical schema with uninformative column names such as \texttt{A2-A16}, making interpretation difficult without external documentation or domain knowledge. The column names provide little semantic meaning, and the accompanying data varying amount of information, complicating their use in text-to-SQL query generation and requiring additional metadata for effective database interactions.}
  \label{fig:schema_example}
\end{figure}

Text-to-SQL, which involves converting natural language queries into Structured Query Language (SQL), holds promise as it facilitates broader access to structured data in SQL databases by reducing the need for expert knowledge. However, working with SQL databases often presents challenges due to ambiguous or uninformative column names, hindering the effectiveness of both human users and generative models employed in text-to-SQL tasks \cite{drspider, chess_stanford}. An illustration of such uninformative column names can be found in the \texttt{district} table in the financial database of the BIRD-Bench text-to-SQL benchmark \cite{BIRD-Bench-NeurIPS}, which is shown in Figure \ref{fig:schema_example}. The columns labeled \texttt{A2-A16} have completely uninformative names and are therefore hard to interpret without either amassed domain knowledge or access to external database documentation.

To address this challenge, recent works have explored using natural language descriptions or metadata as a semantic layer atop existing database structures, significantly enhancing their usability \cite{chess_stanford, ta-sql, BIRD-Bench-NeurIPS}. The critical role of such metadata is further underscored by the fact that all of the top 20 models on the BIRD-Bench benchmark incorporate by the benchmark-provided metadata to attain their high scores, as evidenced by the \textit{Oracle Knowledge} checkbox on the benchmark's website\footnote{https://bird-bench.github.io/}. However, this metadata was manually crafted, making it both time-consuming and labor-intensive to produce. Given the widespread use of relational databases across society, each varying in structure and size, there is a pressing need for automated systems capable of generating natural language descriptions for SQL databases to enhance their interpretability for both human users and text-to-SQL models.

A related area to generating column descriptions is semantic column type detection, where models infer a column's data type based on its content or schema \cite{annotating_columns, semantic_type_detection_sato}. However, these methods typically provide only basic type annotations, lacking the context needed to fully understand the column's purpose. For example, identifying a column as "dates" doesn't clarify whether they are transaction or birth dates. This limited detail leaves users and text-to-sql models to interpret the column's exact meaning. What’s needed are systems that generate detailed, human-readable descriptions of the column’s meaning.

The recent advancements in LLMs have shown that they can produce coherent and contextually relevant text across various domains \cite{llms-few-shot-learners}, making them potential tools for automating such metadata creation. In the context of SQL databases, the hypothesis is that LLMs could leverage multiple signals from a database to generate meaningful column descriptions, including the schema structure, relationships between tables, and the values present within the database itself. Previous research has used generative LLMs to expand abbreviated column names in a SQL setting \cite{zhang-etal-2023-nameguess}, but to the best of our knowledge, no works exist for generating more detailed descriptions across a wider array of column types and formats.

Motivated by these observations, we therefore investigate the capability of LLMs to generate column descriptions that are both detailed and contextually relevant, and how such metadata can enhance the effectiveness and robustness of text-to-SQL models. To enable this study, we first construct a dataset specifically for database description generation using a mix of manual and automated techniques to improve the existing column descriptions provided by the BIRD-Bench benchmark. Second, expert human annotators annotate the difficulty of generating a column description for each specific column to provide a deeper understanding of what table contexts are possible for an LLM to understand. 

We then evaluate several instruction-tuned models for column description generation and assess the impact of different types of column descriptions on text-to-SQL model performance. Our findings reveal that models such as Mixtral 22Bx8 and Command R+ excel in generating high-quality descriptions. However, generating accurate descriptions for columns with inherent ambiguity remains a challenge. Despite this, our experiments show that including detailed column descriptions, whether manually refined or synthetically generated, consistently enhances text-to-SQL accuracy across a range of LLMs.

In this paper, we make the following contributions:

\begin{itemize}
    \item We create a new dataset of column descriptions based on the BIRD-Bench benchmark, improving the quality of existing column descriptions through a combination of LLM-generated outputs and manual refinements. 
    \item We introduce a difficulty annotation scheme to classify columns based on the ease with which meaningful descriptions can be generated, offering insights into the inherent challenges faced by models.
    \item We evaluate several state-of-the-art LLMs, including GPT-4o, Command R+, Qwen2 72B, and others, in their ability to generate detailed and contextually accurate column descriptions, assessing their performance against the gold standard.
    \item We conduct a series of text-to-SQL evaluations to analyze the effect of different types of column descriptions on SQL query generation, revealing that more detailed metadata, even if perceived as superfluous by humans, can significantly improve model performance.
    \item Finally, we propose a two-step process for handling inherently ambiguous columns, combining automated generation with human expert intervention. We also highlight key areas for future research, including exploring additional metadata types to enhance text-to-SQL systems further.
\end{itemize}

\section{Related Work}

\subsection{Table Understanding and Metadata Augmentation Tasks}

In recent years, significant effort has been devoted to simplifying the understanding of tabular data. Table-to-text \cite{investigatin-table2text-yale, qtsumm-yale-harvard, kasner-etal-2023-tabgenie, yang-etal-2022-tableformer, gong-etal-2019-table} and table question answering \cite{pal-etal-2023-multitabqa, xie-etal-2022-unifiedskg, herzig-etal-2020-tapas} aim at developing models able to understand structured tabular data and natural language questions to perform reasoning and tasks across tables.

Previous studies on column type detection and relationship identification have aimed to enhance table understanding by classifying columns into semantic categories (e.g., dates, currencies) \cite{annotating_columns, semantic_type_detection_sato} and identifying structural relationships (e.g., primary and foreign keys) \cite{column_relation_tcn, column_relation_turl}. In contrast, our approach uses LLMs to generate natural language descriptions, providing a more comprehensive and contextually rich understanding of column contents.

While the work of Zhang et al.~\cite{zhang-etal-2023-nameguess} focuses on expanding abbreviated column names to improve database schema readability and table understanding tasks using LLMs, our work goes further by generating full descriptions for column contents and considering a much more diverse range of column names.

\subsection{Metadata Usage in Text-to-SQL}

The shift from traditional query languages to natural language interfaces through text-to-SQL has been significantly advanced by current LLMs. However, the inherent ambiguities and missing information in database schemas necessitate the incorporation of external knowledge and metadata to improve conversion accuracy \cite{NLforTabular_journal}. Previous studies have explored various approaches to address this challenge. Li et al. \cite{chess_stanford} propose a pipeline that retrieves relevant data and context from a knowledge store to generate accurate SQL queries, leveraging a data store based on the original BIRD-Bench metadata files. Similarly, Dou et al. \cite{dou-etal-2022-towards-knowledge-intensive} enhance the text-to-SQL process by using external domain-specific knowledge, such as mathematical formulas and publicly available resources.

Sun et al. \cite{sun2024sqlpalm} improve tuning performance using question-specific database content, while Hong et al. \cite{hong2024knowledgetosql} employ a data expert LLM to aid SQL generation. Petrovski et al. \cite{resilient-sql-petrovski} take a different approach by embedding entire data columns, removing the need for traditional metadata. These studies demonstrate the value of external knowledge and metadata in enhancing text-to-SQL systems.

In contrast to these approaches, our work focuses on generating comprehensive SQL column descriptions directly using LLMs, rather than using humanly curated descriptions.

\section{Column Description Dataset Creation}

To enable our study on generating informative column descriptions, we create a new dataset based on the development set of the widely-used text-to-SQL benchmark, BIRD-Bench \cite{BIRD-Bench-NeurIPS}\footnote{BIRD-Bench is licensed under CC BY-SA 4.0, permitting transformation and reuse of the dataset.}. This dataset consists of 799 columns across 11 different databases and domains, which are presented in Table \ref{tab:dataset_composition}. We selected BIRD-Bench as the basis for our dataset because it allows us to test column description generation across a broad set of domains. Additionally, the natural language queries in BIRD-Bench were designed so that external knowledge (e.g., metadata) is necessary to perform well on the task. In contrast, the WikiSQL \cite{zhong2017seq2sql} and SPIDER \cite{yu-etal-2018-spider} text-to-SQL benchmarks primarily uses exact column names in its natural language queries, making them unsuitable for our use case.

While BIRD-Bench already includes column descriptions, many lacked sufficient detail or contained inaccuracies. To address this, we improved the existing descriptions using a combination of automated LLM and manual refinements, which allowed us to create a gold standard for evaluating model performance. Note that we chose to focus on generating column descriptions for this work, and leave the generation of descriptions of values inside the database for future work.

\subsection{Improving Existing Column Descriptions with LLMs}
\label{sec:improving-bird}

We aimed to improve existing column descriptions using LLMs both as a preliminary step before human annotators created the gold standard dataset, and to evaluate the effectiveness of LLMs in improving descriptions.

To improve existing column descriptions, we provided GPT-4o the BIRD-Bench column descriptions, corresponding schema, example data and an instruction to generate new descriptions. The full prompt template can be found in Appendix \ref{app:prompt}. The original BIRD-Bench descriptions and the generated ones were then independently assessed by two human annotators experts in SQL. Descriptions were categorized as either "Perfect," "Somewhat Correct," "Incorrect," "No Description," or "I can't tell." The full annotator guidelines and more detailed descriptions of the labels are presented in Apendix \ref{app:anno-quality-improved}.

The results show that the original BIRD-Bench descriptions were "Perfect" for 40.24\% of columns, while LLM-generated descriptions improved this to 70.80\%. The percentage of "Somewhat Correct" descriptions decreased from 38.80\% to 18.80\% after LLM improvement, and errors were nearly eliminated (dropping from 1.88\% to 0.06\%). A table showcasing the full results of the annotations  can be found in Appendix \ref{app:improve-column-descriptions}. 

Cohen’s kappa indicated substantial agreement between annotators, with scores of 0.68 for the original descriptions and 0.61 for the LLM-generated ones. While not perfect, LLM-generated descriptions significantly improved over the original dataset and proved useful for improving the quality of column descriptions with existing issues. Full details of the results are presented in Appendix \ref{app:improve-column-descriptions}.

\subsection{Creating the Gold Standard Dataset}
\label{sec:benchmark}

Since not all LLM-generated improvements were perfect, manual corrections of the generations were necessary to achieve the high standard required for a gold standard dataset. First, disagreements in scores were addressed by reviewing each other's annotations and then through discussion to reach a consensus. Descriptions that did not achieve a ``Perfect'' quality score from both annotators after this phase were manually corrected until both annotators agreed on a ``Perfect'' rating. The final set of descriptions was then reviewed to ensure they provided clear, comprehensive, and contextually rich information for each column.
Through this process, we believe we have created a high quality dataset serving as a reliable reference for evaluating the performance of models in generating column descriptions. The final composition of the dataset can be found in Table \ref{tab:dataset_composition}. Examples of gold descriptions can be found in Appendix \ref{app:column_examples}.

\begin{table*}[t]
\centering
\small
\begin{tabularx}{\textwidth}{lCCCCCC}
\toprule
\textbf{Database} & \textbf{\# tables} & \textbf{\# columns} & \textbf{Self-Evident} & \textbf{Context-Aided} & \textbf{Ambiguity-Prone} & \textbf{Domain-Dependent} \\
\midrule
debit\_card\_specializing & 5 & 21 & 17 & 3 & 1 & 0 \\
financial & 8 & 55 & 21 & 9 & 11 & 13 \\
formula\_1 & 13 & 94 & 49 & 35 & 10 & 0 \\
california\_schools & 3 & 89 & 24 & 32 & 30 & 3 \\
card\_games & 6 & 115 & 37 & 28 & 50 & 0 \\
european\_football\_2 & 7 & 199 & 34 & 62 & 103 & 0 \\
thrombosis\_prediction & 3 & 64 & 56 & 2 & 6 & 0 \\
toxicology & 4 & 11 & 8 & 0 & 2 & 1 \\
student\_club & 8 & 48 & 39 & 9 & 0 & 0 \\
superhero & 10 & 31 & 30 & 1 & 0 & 0 \\
codebase\_community & 8 & 71 & 62 & 4 & 5 & 0 \\
\midrule
\textbf{Total} & 75 & 798 & 377 (47\%) & 185 (23\%) & 218 (27\%) & 17 (2\%) \\
\bottomrule
\end{tabularx}
\caption{Number of tables and columns, and the distribution of column difficulty across the databases in our column description dataset.}
\label{tab:dataset_composition}
\end{table*}

\subsection{Annotating Column Difficulty}
\label{sec:column-difficulty}
After finalizing the gold standard descriptions, columns were categorized into four difficulty levels: "Self-Evident," "Context-Aided," "Ambiguity-Prone," and "Domain-Dependent," based on the amount of information available in the database to generate accurate descriptions.

For "Self-Evident" columns, like \texttt{birth\_date} in the \texttt{Client} table, the schema alone was sufficient to generate an accurate description ie. the column's meaning was self-evident. "Context-Aided" columns, such as \texttt{short\_state} in the \texttt{zip\_code} table, required signals from both the schema and the data for accurate interpretation. This suggests that while some additional context is required, the column's role can still be deduced by looking at the data inside the database. "Ambiguity-Prone" columns posed challenges even with the available schema and data due to uncertainties; for example, in the \texttt{formula\_1} database's \texttt{results} table, the \texttt{rank} column has data indicating rankings, but without context, it is hard to determine if it refers to race position, qualifying rank, or another metric. The description, "The overall rank of the driver's fastest lap time in the race," clarifies its purpose. "Domain-Dependent" columns, like \texttt{A11} in the \texttt{districts} table, were impossible to describe without access to additional documentation or domain knowledge.

Two independent annotators, separate from those who created the gold descriptions, assessed the difficulty, achieving substaintial agreement (Cohen’s kappa of 0.61). Disagreements were resolved through discussion, resulting in a final difficulty score for each column. Most columns fell into the "Self-Evident" and "Context-Aided" categories, while "Domain-Dependent" columns were primarily found in the Financial domain.

We believe that the difficulty ratings of columns will aid in understanding and improving the performance of models in generating column descriptions across varying levels of complexity.

\section{Experiments}

\subsection{Generating Column Descriptions}
\label{sec:model-eval}

To evaluate the effectiveness of LLMs in generating column descriptions, we used our dataset presented in Section \ref{sec:benchmark}. We evaluated the following models: GPT-4o, Mixtral-8x22B-Instruct-v0.1 \cite{jiang2024mixtral}, Command R+, Qwen2-72B-Instruct \cite{qwen2}, Codestral-22B-v0.1, and Mistral-7B-Instruct-v0.3 \cite{jiang2023mistral}. These models were selected to provide a diverse range of perspectives, including a state-of-the-art closed-source model (GPT-4o), several newly released high-performing open-source models (Qwen2-72B, Mixtral-8x22B, Codestral-22B, Command R+), and a smaller model (Mistral-7B).

Each model was given the database schema, the specified column, example data rows from the corresponding table, and specific instructions for generating the column description. The exact prompt is presented in Appendix \ref{app:prompt}. 

Inference for the open-source models was performed on several A100 GPUs, using BitsandBytes \cite{dettmers2022llm} to load the models in 8-bit format. We set the model temperature to 0.7.

\subsection{Identifying Ambiguous Columns}
\label{sec:id-ambiguous-columns}

For SQL databases, some columns might be inherently ambiguous as described in Section \ref{sec:column-difficulty}, making it difficult to generate accurate descriptions without domain knowledge. To address this, we conducted a small ablation study to evaluate whether LLMs could autonomously recognize the "Ambiguity-Prone" or "Domain-Dependent" columns—specifically those that exhibit inherent ambiguity. The goal was to assess the model's ability to detect situations where the schema and data alone are insufficient for making unambiguous predictions about the columns' contexts.

In our experiment, we modified the prompt provided to the LLMs, including the instruction: "If the details in the schema do not suffice to ascertain what the columns meaning is without ambiguity, return: 'Not enough information to make a valid prediction.'" This addition was designed to prompt the LLM to identify and flag ambiguous columns rather than generating potentially misleading descriptions.

By leveraging this capability, we propose a two-step approach: first, automatically generating descriptions for columns the LLMs identify as unambiguous, and second, involving human experts to annotate the remaining columns where inherent ambiguity is detected. 

\subsection{Text-to-SQL Evaluations}

\begin{figure*}[t]
    \centering    
    \includegraphics[scale=0.37, trim={0.25cm 0 0 0}, clip]{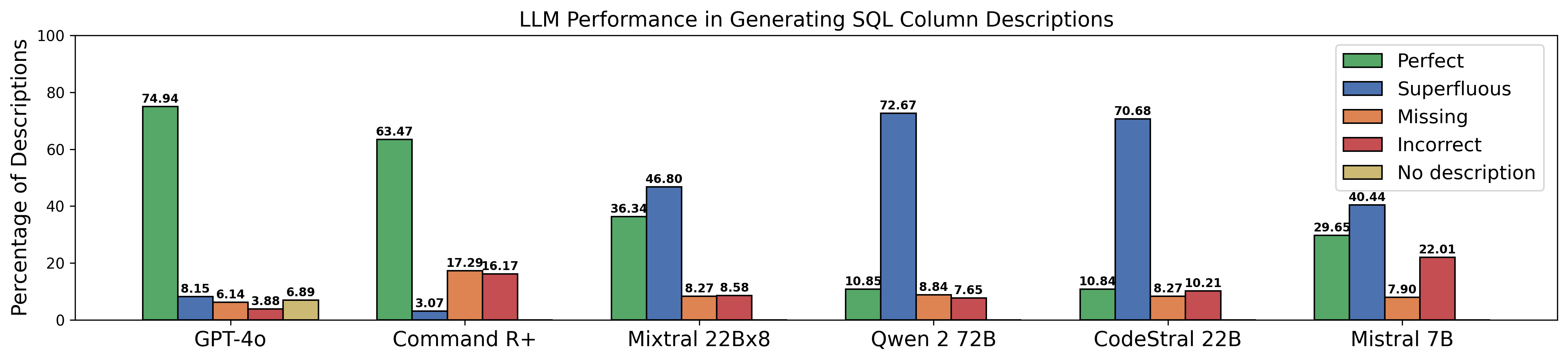}
    \caption{Visual representation of LLM performance in generating SQL column descriptions.}
    \label{fig:model_quality_bar_plot}
\end{figure*}

\begin{table*}[t]
\centering
\small
\begin{tabularx}{\textwidth}{lYYYYYY}
\toprule
\textbf{Difficulty} & \textbf{GPT-4o} & \textbf{Command R+} & \textbf{Mixtral 22x8B} & \textbf{Qwen 2 72B} & \textbf{CodeStral 22B} & \textbf{Mistral 7B} \\
\midrule
Self-Evident       & 3.83 & 3.55 & 3.26 & 2.98 & 2.97 & 3.24 \\
Context-Aided     & 3.66 & 3.28 & 3.18 & 2.89 & 2.79 & 2.84 \\
Ambiguity-Prone       & 3.46 & 2.46 & 2.90 & 2.76 & 2.69 & 2.04 \\
Domain-Dependent  & 1.93 & 1.24 & 1.59 & 1.47 & 1.44 & 1.12 \\
\bottomrule
\end{tabularx}
\caption{Mean quality scores for each model's generated descriptions across the column difficulty levels in the dataset, as rated by the human evaluators.}
\label{tab:mean_quality_difficulties}
\normalsize
\end{table*}

To study the usefulness of column descriptions for text-to-SQL systems, we first evaluated the same LLMs as in the description generation step in a zero-shot setting on the development set of BIRD-Bench, which contains 1534 natural language questions and SQL query pairs. Our goal was to isolate and understand the effects of column descriptions on model performance.

The experimental setup involved five scenarios: (1) Without any column descriptions, where the model operates with only the database schema; (2) With BIRD-Bench original column descriptions, utilizing the original descriptions provided by BIRD-Bench in addition to the schema; (3) With gold standard descriptions, where the model uses our manually refined gold standard descriptions in addition to the schema; (4) With GPT-4o generated column descriptions, where the model uses descriptions generated by the best model from Section \ref{sec:model-eval} in addition to the schema; and (5) With Qwen2 generated descriptions, where the model uses descriptions rated with a high amount of Superflous information. We included both the GPT-4o and the Qwen2-generated descriptions to test whether our humanly defined categories of "Perfect" and "Superfluous information" are aligned with what is practically useful for LLMs in the text-to-SQL task. In all scenarios, we included only column descriptions as the available metadata alongside the database schema. We set the model temperature to 0. The full prompt for the text-to-SQL generation is presented in Appendix \ref{app:prompt}.

\section{Results}

\subsection{Human Evaluation of Generated Column Descriptions}

For each model's generated output, two human annotators independently assessed the quality of the column descriptions. As previously the set of annotators was different from the ones that created the gold dataset, but with the same type of expertise. The annotations were also based on a slightly different set of criteria from what was used when rating the original and improved column descriptions, presented in Section \ref{sec:improving-bird}, as we now have the gold column descriptions as a reference point.

The generated outputs were scored for their usefulness, using labels (4) ``Perfect'', (3) ``Superfluous Information'', (2) ``Missing Information'', (1) ``Incorrect'',  and ``No Description''. The main difference between ``Perfect'' and ``Superfluous Information'', is that the latter contains additional redundant information in addition to the gold standard that is not incorrect but deemed excessive by the annotators. A ``Missing information'' column description has partly correct information, but information is missing compared to the gold standard. The full instructions and definitions of the labels are presented in Appendix \ref{app:anno-quality-pred}.

To ensure the reliability of our evaluations, we measured the inter-annotator agreement for each model’s output. The agreement among annotators was above 0.60 for all models, indicating a substantial agreement between the annotators.

The results of the model evaluations are presented in Figure \ref{fig:model_quality_bar_plot} and a breakdown of performance over the different difficulty levels can be found in Table \ref{tab:mean_quality_difficulties}. Exact numbers of the LLM performances can be found in Appendix \ref{app:exact_performance}. GPT-4o demonstrates superior capability, producing perfect descriptions for 74.94\% of the columns, substantially outperforming other models in this category. However, it's important to note that the gold dataset was created using GPT-4o, which introduces a bias in favor of this model's performance. As such, the GPT-4o results are included primarily for comparison purposes and should be interpreted cautiously. Command R+ follows with 63.47\% perfect descriptions, though it also shows notable percentages of missing (17.29\%) and incorrect (16.17\%) information. Mixtral 22x8B, Qwen 2 72B, and CodeStral 22B strongly favor providing redundant information, with percentages ranging from 46.80\% to 72.67\%. This suggests these models may be over-generating or including unnecessary details in their descriptions. Mistral 7B presents a more balanced distribution across categories, with 29.65\% perfect descriptions and 40.44\% superfluous information, indicating a moderate performance level.

Table \ref{tab:mean_quality_difficulties} highlights model performance across column difficulty levels. GPT-4o leads in all categories, scoring 3.83 for "Self-Evident" and 3.66 for "Context-Aided" columns, while other models like Command R+ and Mixtral 22x8B perform slightly lower. However, GPT-4o's scores are as mentioned likely biased, as the dataset was partly created using GPT-4o itself.

As difficulty increases with the "Ambiguity-Prone" and "Domain-Dependent" columns, all models see a drop in performance. GPT-4o scores 3.46 and 1.93, respectively, but struggles alongside the other models in these categories, with Mistral 7B showing the weakest performance. These results underscore the need for better handling of complex and ambiguous columns across all models, and indicate the need for further documentation or domain knowledge.

GPT-4o was the only model to produce "No description" outputs (6.89\%), abstaining when insufficient information was available, unlike other models. The difficulty split for these columns shows GPT-4o flagged 17 as "Self-Evident," 8 as "Context-Aided," 26 as "Ambiguity-Prone" and 3 as "Domain-Dependent." While it successfully identified many ambiguous columns, it also flagged some that were straightforward. Its flagging of easier column difficulties indicates the need for better ways of handling and dealing with ambiguous columns than our approach.

We also explored using GPT-4o as an automated judge to evaluate the quality of generated column descriptions. However, the results showed low alignment with human evaluations, with Cohen’s Kappa scores ranging from 0.12 to 0.25, indicating that LLM-as-a-judge was not a reliable substitute for human judgment. We included the results in Appendix \ref{app:llm-judge-eval} and the prompt used in Appendix \ref{app:prompt} for future work to explore.

\subsection{Impact of Column Descriptions on Text-to-SQL Performance}

\begin{table*}[t]
\centering
\small
\begin{tabularx}{\textwidth}{lYYYYYY}
\toprule
\textbf{Included Descriptions} & \textbf{Command R+} & \textbf{Mistral 7B} & \textbf{CodeStral 22B} & \textbf{Qwen2 72B} & \textbf{GPT-4o} & \textbf{Mixtral 22Bx8} \\
\midrule
No descriptions       & 0.2321 & 0.1473 & 0.2529 & 0.2894 & 0.3013 & 0.1750\\
BIRD-Bench descriptions     & 0.2419 & 0.1506 & 0.2647 & 0.3116 & 0.3134 & 0.2543 \\
Gold descriptions     & 0.2516 & 0.1649 & 0.2738 & 0.3220 & 0.3328 & 0.2716 \\
GPT-4o generated descriptions & 0.2445 & 0.1538 & 0.2637 & 0.2602 & 0.3206 & 0.2583 \\
Qwen2 generated descriptions & \textbf{0.2675} & \textbf{0.1690} & \textbf{0.2966} & \textbf{0.3349} & \textbf{0.3678} & \textbf{0.2959} \\
\bottomrule
\end{tabularx}
\caption{Zero-shot execution accuracy scores for different types of metadata across various LLMs.}
\label{tab:model_scores_metadata}
\normalsize
\end{table*}

The full results of our text-to-SQL experiments are presented in Table \ref{tab:model_scores_metadata}. The text-to-SQL experiments reveal that including column descriptions generally improves the performance of all tested models, with the largest gains seen when using descriptions generated by Qwen2. Across the board, Qwen2-generated descriptions consistently outperformed both the gold descriptions and GPT-4o-generated descriptions, suggesting that the additional, seemingly superfluous information in Qwen2's descriptions is actually beneficial for LLMs. This is a surprising result, as it contradicts the annotators' initial assessment of what constitutes useful versus redundant information.

While all models demonstrated better accuracy with Qwen2 descriptions, the largest models, such as GPT-4o and Qwen2 72B, showed the most significant performance increases. For example, GPT-4o’s accuracy rose from 0.3013 without descriptions to 0.3678 with Qwen2 descriptions, indicating that larger models, with their greater capacity to understand semantics and context, may better leverage detailed metadata. In contrast, smaller models like Mistral 7B saw more modest gains, with accuracy improving by just 2\% when descriptions were included (from 0.1473 without descriptions to 0.1690 with Qwen2 descriptions). 

In addition to the strong performance of the synthetically generated descriptions, our manually crafted gold descriptions also outperformed the humanly annotated BIRD-Bench descriptions across all models. This indicates that more carefully curated descriptions, whether humanly refined or synthetically generated, provide greater clarity and utility for the models.

To further explore the impact of column descriptions, an ablation study detailed in Appendix \ref{app:difficult-columns-ablation} shows that models gained over 20\% in accuracy when provided with descriptions for completely uninformative columns, further underscoring the important role of detailed metadata in enhancing text-to-SQL performance.

\section{Conclusions and Future Work}

In this work, we demonstrated that natural language descriptions and metadata are crucial for improving the performance of text-to-SQL models. Our experiments show that including detailed column descriptions, whether manually refined or synthetically generated, consistently enhances query generation accuracy across a range of LLMs. However, generating accurate descriptions for columns with inherent ambiguity unsurprisingly remains a challenge. For handling such columns we propose a two-step approach: first, automatically generate descriptions for columns with clear interpretation signals in the database; then, have a domain expert manually describe inherently ambiguous columns. This ensures LLMs handle straightforward columns independently while relying on expert input for ambiguous cases, though further research is needed to pinpoint these columns.

To address this, we propose a two-step approach for handling ambiguous columns: first, automatically identify and generate descriptions for columns where meaningful interpretation signals exist within the database. Then, for columns with inherent ambiguity, a domain expert should manually craft the necessary descriptions. This strategy ensures that the LLMs can handle straightforward columns autonomously while relying on expert input where ambiguity is present, although more research is needed to identify such ambiguous columns.

Our findings also suggest that models interpret and utilize detailed information differently than humans might expect. Specifically, descriptions that were initially categorized as containing "superfluous" information led to improved query generation performance across all models. This highlights the potential value of including more detailed metadata than humans might consider necessary.

Future work will focus on understanding which specific features of the Qwen2-generated descriptions contributed to the significant performance gains seen across all models. Additionally, we aim to explore what types of information are most beneficial for improving text-to-SQL performance, particularly in the context of ambiguous columns. Finally, future research will investigate the incorporation of other metadata types, such as value descriptions, synonyms, and numerical reasoning, to further enhance the capabilities of text-to-SQL systems.

\section*{Limitations}

In this study, we restricted our evaluation to the development set of BIRD-Bench due to the labor costs associated with performing the multitude of human evaluations required. Although the full training set encompasses more domains and might offer a broader evaluation scope, we believe that the development set contains sufficient variety to ensure robust testing of the models. The chosen subset still represents a wide range of domains, allowing us to draw meaningful conclusions about the performance of large language models in generating column descriptions.

Another limitation is that our annotators were not experts in all the specific domains covered by the dataset, such as Thrombosis Prediction. Despite this, the annotators reported confidence in their ability to accurately label column descriptions across all domains, including these more challenging areas. Nevertheless, the lack of domain-specific expertise could introduce some bias or inaccuracies in the annotation process. Future work should consider incorporating domain experts for annotation, particularly for specialized fields, to further enhance the dataset's reliability.

\section*{Ethical Statement}

All annotators involved in this study were fully informed about the workload and had the option to decide whether they wanted to participate. Since all annotators were also authors of this paper, they were not compensated for their efforts. However, we ensured that the annotations were produced ethically, with voluntary participation and informed consent from all contributors.

It is important to note that the metadata generation models presented in this paper are not perfect. A model that achieves high scores on our dataset may not generalize well to scenarios outside of the domains we studied. Consequently, caution should be exercised when using these models to generate column descriptions in systems that will be used by humans or other AI systems, such as text-to-SQL models. Ensuring the reliability and accuracy of generated metadata in different contexts is crucial to prevent potential misinterpretations or errors in downstream applications. 

\section*{Acknowledgements}

The computations in this paper was enabled by the Berzelius resource provided by the Knut and Alice Wallenberg Foundation at the National Supercomputer Centre in Linköping, Sweden, to whom we extend our gratitude. Additionally, this research was also partially funded by ELLIIT, a strategic research environment in information technology and mobile communications funded by the Swedish government.

\newpage

\bibliographystyle{unsrt}
\bibliography{references}

\appendix
\onecolumn

\section{Datasheet for Datasets}
\label{sec}

\subsection{Motivation}

\paragraph{For what purpose was the dataset created?}

Relational databases often suffer from uninformative descriptors of table contents, such as columns and values, impacting both human users and text-to-SQL models. This dataset was created to explore and assess how large language models can generate detailed and meaningful column descriptions to be used as a semantic layer over databases, to enhance their usability.

\paragraph{Who created this dataset (e.g., which team, research group) and on behalf of which entity (e.g., company, institution, organization)?}

This question will be answered after the anonymity period.

\paragraph{What support was needed to make this dataset?}

This question will be answered after the anonymity period.

\subsection{Composition}

\paragraph{What do the instances that comprise the dataset represent (e.g., documents, photos, people, countries)?}

The dataset is made up of 11 SQLite databases, accompanied by a csv file which contains gold standard column descriptions and difficulty ratings for every column in the entire dataset. Each instance therefore represents a column and its accompanying gold description and difficulty rating.

\paragraph{How many instances are there in total (of each type, if appropriate)?}

There are 798 columns in total across the 11 databases in the dataset. A full breakdown of columns over the different domains can be found in Table \ref{tab:dataset_composition}.

\paragraph{Does the dataset contain all possible instances or is it a sample (not necessarily random) of instances from a larger set?}

The dataset is based on the BIRD-Bench development set and includes all databases, tables, and columns from the development set.

\paragraph{What data does each instance consist of?}

Each instance contains a column, its corresponding table and database, and its accompanying gold description and difficulty rating.

\paragraph{Is there a label or target associated with each instance? If so, please provide a description.}

Yes. Each column contains a corresponding gold column description, as well as an annotated difficulty rating (Easy, Medium, Hard or Very Hard). These levels reflect the annotators' assessment of the amount of available information necessary to generate an accurate description for a column. Examples of columns of each difficulty can be found in \ref{app:column_examples}.

\paragraph{Is any information missing from individual instances?}

No. The dataset has full coverage over the entire set of columns from the included databases.

\paragraph{Are there any errors, sources of noise, or redundancies in the dataset?}

The dataset was created using a combination of automated techniques and human annotators. Although the goal was to create a dataset free of noise, it could be that some noise exists inadvertently due to human error.

\paragraph{Is the dataset self-contained, or does it link to or otherwise rely on external resources (e.g., websites, tweets, other datasets)?}

The dataset is self-contained.

\paragraph{Does the dataset contain data that might be considered confidential (e.g., data that is protected by legal privilege or by doctor-patient confidentiality, data that includes the content of individuals’ non-public communications)?}

No.

\paragraph{Does the dataset contain data that, if viewed directly, might be offensive, insulting, threatening, or might otherwise cause anxiety?}

No.

\subsection{Collection}

\paragraph{How was the data associated with each instance acquired? Was the data directly observable (e.g., raw text, movie ratings), reported by subjects (e.g., survey responses), or indirectly inferred/derived from other data (e.g., part-of-speech tags, model-based guesses for age or language)? If data was reported by subjects or indirectly inferred/derived from other data, was the data validated/verified? If so, please describe how.}

The data was directly observable and derived from the BIRD-Bench development set. The column descriptions were generated using a large language model and then manually verified and corrected by human annotators. The difficulty ratings where annotated and verified by human annotators. 

\paragraph{Over what timeframe was the data collected? Does this timeframe match the creation timeframe of the data associated with the instances (e.g., recent crawl of old news articles)? If not, please describe the timeframe in which the data associated with the instances was created. Finally, list when the dataset was first published.}

The data was collected and the column descriptions were generated and annotated over a period of three months. The dataset will be published after the anonymity period.

\paragraph{What mechanisms or procedures were used to collect the data (e.g., hardware apparatus or sensor, manual human curation, software program, software API)? How were these mechanisms or procedures validated?}

The dataset was collected from the BIRD-Bench development set. LLMs were used to generate initial column descriptions, which were then manually curated and corrected by human annotators. The validation was performed by expert annotators through a consensus process.

\paragraph{What was the resource cost of collecting the data? (e.g., what were the required computational resources, and the associated financial costs, and energy consumption - estimate the carbon footprint.}

The resource cost included the computational expense of running LLMs on multiple A100 GPUs and the human labor for manual annotation. The exact financial cost and carbon footprint would require further detailed calculations, but they are comparable to typical machine learning model training and annotation tasks.

\paragraph{If the dataset is a sample from a larger set, what was the sampling strategy (e.g., deterministic, probabilistic with specific sampling probabilities)?}

The dataset is not a sample; it includes the complete BIRD-Bench development set.

\paragraph{Who was involved in the data collection process (e.g., students, crowdworkers, contractors) and how were they compensated (e.g., how much were crowdworkers paid)?}

The data collection and annotation were performed by the authors of the paper, who are researchers with expertise in SQL and natural language processing. There was no additional compensation as the work was part of their research activities.

\paragraph{Were any ethical review processes conducted (e.g., by an institutional review board)? If so, please provide a description of these review processes, including the outcomes, as well as a link or other access point to any supporting documentation.}

No formal ethical review was conducted as the data does not involve personal or sensitive information.

\paragraph{Does the dataset relate to people? If not, you may skip the remainder of the questions in this section.}

No.

\subsection{Preprocessing / Cleaning / Labeling}

\paragraph{Was any preprocessing/cleaning/labeling of the data done (e.g., discretization or bucketing, tokenization, part-of-speech tagging, SIFT feature extraction, removal of instances, processing of missing values)? If so, please provide a description. If not, you may skip the remainder of the questions in this section.}

Yes, preprocessing included several steps:

\begin{itemize}
    \item Data cleaning steps were applied to the original BIRD-Bench databases and dataset:
    \begin{itemize}
        \item Removed non "utf-8" tokens from the original BIRD-bench metadata CSV files.
        \item Corrected spelling in the \texttt{european\_football\_2} database in the "Country" description header from "desription" to "description".
        \item Removed all columns with zero data and the name "Unnamed" due to (too many/missing) commas in the original BIRD-Bench metadata CSV files.
        \item Changed the file name of \texttt{"ruling.csv"} to \texttt{"rulings.csv"} to match the original table name in the \texttt{card\_games} domain.
        \item Changed \texttt{"set\_transactions.csv"} to \texttt{"set\_translations.csv"} to match the database name in the \texttt{card\_games} domain.
        \item Removed all \texttt{.DS\_STORE} files from the data directory.
        \item Changed the names of the CSV files in the \texttt{student\_club} database to match the original table names. Changed upper case to lower case on the first letter of all CSV files, and corrected the code in \texttt{Zip\_Code}.
        \item Removed the "wins" column from \texttt{constructors.csv} file as the column does not exist in the \texttt{formula\_1} database.
    \end{itemize}
    \item Generating initial column descriptions using LLMs, followed by manual correction and labeling of the descriptions.
    \item Categorizing the columns based on the difficulty of generating their descriptions.
\end{itemize}

\paragraph{Was the “raw” data saved in addition to the preprocessed/cleaned/labeled data (e.g., to support unanticipated future uses)? If so, please provide a link or other access point to the “raw” data.}

Yes, the raw BIRD-Bench development set data was saved and is available upon request.

\paragraph{Is the software used to preprocess/clean/label the instances available? If so, please provide a link or other access point.}

Yes, the code used for preprocessing, cleaning, and labeling is available at \url{https://github.com/anonymous} (to be updated post anonymity period).

\subsection{Uses}

\paragraph{Has the dataset been used for any tasks already? If so, please provide a description.}

The dataset has been used to evaluate the performance of various large language models in generating column descriptions and to study the impact of these descriptions on text-to-SQL systems.

\paragraph{Is there a repository that links to any or all papers or systems that use the dataset? If so, please provide a link or other access point.}

This will be updated after the anonymity period with the appropriate links.

\paragraph{What (other) tasks could the dataset be used for?}

The dataset could be used for tasks such as evaluating and improving text-to-SQL models, studying the impact of metadata on database usability, and developing automated systems for generating database documentation.

\paragraph{Is there anything about the composition of the dataset or the way it was collected and preprocessed/cleaned/labeled that might impact future uses? For example, is there anything that a future user might need to know to avoid uses that could result in unfair treatment of individuals or groups (e.g., stereotyping, quality of service issues) or other undesirable harms (e.g., financial harms, legal risks)? If so, please provide a description. Is there anything a future user could do to mitigate these undesirable harms?}

The dataset is composed of data generated and annotated for research purposes and does not include personal or sensitive information. Future users should ensure that the generated metadata is appropriate for their specific use case to avoid misinterpretations or errors in applications.

\paragraph{Are there tasks for which the dataset should not be used? If so, please provide a description.}

The dataset should not be used for applications involving sensitive or personal data without proper validation and consideration of privacy implications.

\subsection{Distribution}

\paragraph{Will the dataset be distributed to third parties outside of the entity (e.g., company, institution, organization) on behalf of which the dataset was created? If so, please provide a description.}

Yes, the dataset will be made publicly available for research purposes.

\paragraph{How will the dataset be distributed (e.g., tarball on website, API, GitHub)? Does the dataset have a digital object identifier (DOI)?}

The dataset will be distributed via GitHub and will be assigned a DOI for reference.

\paragraph{When will the dataset be distributed?}

The dataset will be distributed after the anonymity period.

\paragraph{Will the dataset be distributed under a copyright or other intellectual property (IP) license, and/or under applicable terms of use (ToU)? If so, please describe this license and/or ToU, and provide a link or other access point to, or otherwise reproduce, any relevant licensing terms or ToU, as well as any fees associated with these restrictions.}

The dataset will be distributed under the Creative Commons Attribution-ShareAlike 4.0 International License (CC BY-SA 4.0).

\paragraph{Have any third parties imposed IP-based or other restrictions on the data associated with the instances? If so, please describe these restrictions, and provide a link or other access point to, or otherwise reproduce, any relevant licensing terms, as well as any fees associated with these restrictions.}

No.

\paragraph{Do any export controls or other regulatory restrictions apply to the dataset or to individual instances? If so, please describe these restrictions, and provide a link or other access point to, or otherwise reproduce, any supporting documentation.}

No.

\subsection{Maintenance}

\paragraph{Who is supporting/hosting/maintaining the dataset?}

This will be answered after the anonymity period.

\paragraph{How can the owner/curator/manager of the dataset be contacted (e.g., email address)?}

This will be answered after the anonymity period.

\paragraph{Is there an erratum? If so, please provide a link or other access point.}

Not at this time.

\paragraph{Will the dataset be updated (e.g., to correct labeling errors, add new instances, delete instances)? If so, please describe how often, by whom, and how updates will be communicated to users (e.g., mailing list, GitHub)?}

The dataset will be updated as necessary to correct any errors or add new instances. Updates will be communicated via the GitHub repository.

\paragraph{If the dataset relates to people, are there applicable limits on the retention of the data associated with the instances (e.g., were individuals in question told that their data would be retained for a fixed period of time and then deleted)? If so, please describe these limits and explain how they will be enforced.}

Not applicable.

\paragraph{Will older versions of the dataset continue to be supported/hosted/maintained? If so, please describe how. If not, please describe how its obsolescence will be communicated to users.}

Older versions of the dataset will continue to be available on GitHub. Any deprecation or updates will be clearly communicated through the repository's documentation.

\paragraph{If others want to extend/augment/build on/contribute to the dataset, is there a mechanism for them to do so? If so, please provide a description. Will these contributions be validated/verified? If so, please describe how. If not, why not? Is there a process for communicating/distributing these contributions to other users? If so, please provide a description.}

Contributions can be made through pull requests on the GitHub repository. Contributions will be reviewed and validated by the dataset maintainers before being accepted and merged. All updates and contributions will be documented in the repository's changelog.

\subsection{Any other comments?}

None at this time.

\newpage
\section{Appendix}
\raggedbottom

\subsection{Ablation Study: Analyzing the Effects of Difficult Columns}
\label{app:difficult-columns-ablation}

\begin{figure}[h]
    \centering
    \includegraphics[scale=0.3]{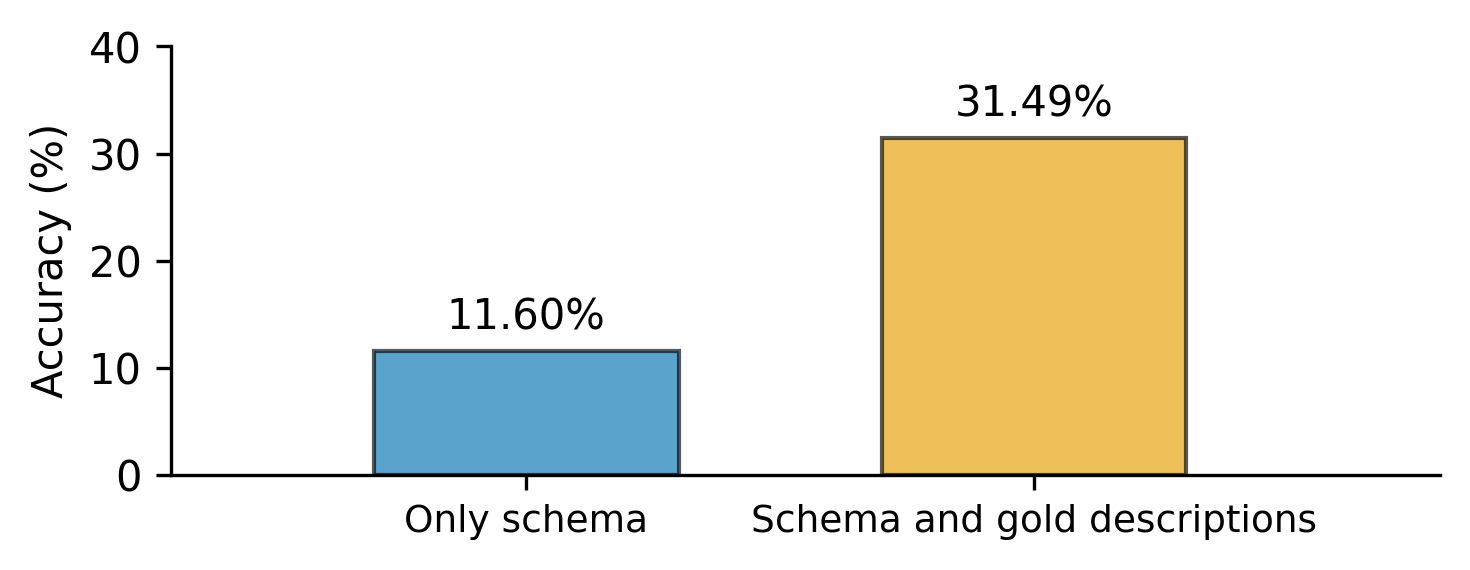}
    \caption{Execution accuracy with and without gold column descriptions on completely uninformative column names.}
    \label{fig:difficult_accuracies}
\end{figure}

To better understand the usefulness of column descriptions in low-information scenarios, we replaced all the column names in each database with uninformative column names. The first column of the first table in database A where denoted by A11, the second column by A12, and so on.

We chose to study this extreme naming scenario to highlight the crucial role of column descriptions when column names provide little to no context. Studying this extreme case is important because it helps us understand the boundaries of model performance and the essential need for descriptive metadata. While such uninformative column names occur in the BIRD-Bench dataset, most realistic scenarios likely fall somewhere between these uninformative names and more informative column names.

The results of this experiment are presented in Figure \ref{fig:difficult_accuracies}. The findings clearly show that column descriptions are highly useful when performing text-to-SQL in scenarios with low-information column names, as the model's performance significantly improves when detailed descriptions are provided.

Through this experiment, we illustrate that detailed column descriptions are essential for the robustness of text-to-SQL systems, especially in challenging scenarios where column names alone are insufficient.

\subsection{Ablation Study: Evaluating Models Using LLM-as-a-Judge}
\label{app:llm-judge-eval}

We performed an evaluation using LLM-as-a-judge, with GPT-4o as the judge. This evaluation aimed to investigate whether there exists an automated metric that could effectively replace human evaluations.

The LLM judge was provided with a task description, the model-generated descriptions, the gold standard references, and the rating criteria. The full prompt is presented in Appendix \ref{app:prompt}.

We compared the model rankings as judged by GPT-4o to those determined by human annotators to see if there was a consistent correlation. The results, as shown in Table \ref{tab:judge-kappa-scores}, indicate low alignment between GPT-4o and human annotators, with Cohen's Kappa scores ranging from 0.12 to 0.25. 

\begin{table}[H]
\centering
\small
\resizebox{1\textwidth}{!}{%
\begin{tabular}{lccccc}
\toprule
\textbf{Annotator/LLM} & \textbf{Command R+} & \textbf{CodeStral 22B} & \textbf{Mistral 7b} & \textbf{Qwen 2 72B} & \textbf{Mixtral 22B} \\
\midrule
Annotator 1 vs GPT-4o & 0.22 & 0.16 & 0.24 & 0.21 & 0.16 \\
Annotator 2 vs GPT-4o & 0.25 & 0.18 & 0.23 & 0.22 & 0.17 \\
\bottomrule
\end{tabular}
}
\caption{Cohen's Kappa scores between annotators and GPT-4o LLM judge for various models.}
\label{tab:judge-kappa-scores}
\end{table}

\subsection{Annotation Scores of the Original BIRD-Bench Descriptions and LLM Improved Descriptions}
\label{app:ann-scores}

\begin{table}[H]
\centering
\small
\begin{tabularx}{0.7\linewidth}{>{\hsize=11\hsize}X >{\hsize=0.5\hsize}c >{\hsize=0.5\hsize}c}
\toprule
\textbf{Descriptions} & \textbf{Agreement Percentage} & \textbf{Cohen's Kappa} \\
\midrule
BIRD-Bench descriptions & 0.79 & 0.68 \\
LLM improved descriptions & 0.82 & 0.61 \\
\bottomrule
\end{tabularx}
\caption{Agreement Percentage and Cohen's Kappa for the quality annotations for both the original BIRD-Bench descriptions and our improved generated descriptions (before corrections).}
\end{table}
\normalsize

\begin{table}[H]
\centering
\small
\begin{tabularx}{0.7\linewidth}{>{\hsize=9\hsize}X >{\hsize=0.5\hsize}c >{\hsize=0.5\hsize}c}
\toprule
\textbf{Difficulty Annotation} & \textbf{Agreement Percentage} & \textbf{Cohen's Kappa} \\
\midrule
Difficulty rating & 0.53 & 0.31 \\
\bottomrule
\end{tabularx}
\caption{Agreement Percentage and Cohen's Kappa for Difficulty Ratings}
\end{table}
\normalsize

\subsection{Annotator Agreement Scores for Generated Column Descriptions}
\label{sec:annotator-eval}

\begin{table}[H]
\centering
\small
\resizebox{0.6\textwidth}{!}{%
\begin{tabular}{lcc}
\toprule
\textbf{Annotator Agreement} & \textbf{Cohen's Kappa} \\
\midrule
Annotator 1 vs annotator 2 (Command R+) & 0.60 \\
Annotator 1 vs annotator 2 (CodeStral 22B) & 0.71 \\
Annotator 1 vs annotator 2 (Mistral 7b) & 0.71 \\
Annotator 1 vs annotator 2 (Qwen 2 72B) & 0.62 \\
Annotator 1 vs annotator 2 (Mixtral 22B) & 0.63 \\
\bottomrule
\end{tabular}
}
\caption{Cohen's Kappa scores between human annotators for various models.}
\label{tab:annotator-kappa-scores}
\end{table}

\subsection{Exact Performance Metrics of the LLM Generated Column Descriptions}
\label{app:exact_performance}

\begin{table*}[h]
\centering
\small
\begin{tabularx}{\textwidth}{lYYYYYY}
\toprule
\textbf{Quality} & \textbf{GPT-4o} & \textbf{Command R+} & \textbf{Mixtral 22x8B} & \textbf{Qwen 2 72B} & \textbf{CodeStral 22B} & \textbf{Mistral 7B} \\
\midrule
Perfect          & 74.94 & 63.47 & 36.34 & 10.85 & 10.84 & 29.65 \\
Superfluous      & 8.15  & 3.07  & 46.80 & 72.67 & 70.68 & 40.44 \\
Missing          & 6.14  & 17.29 & 8.27  & 8.84  & 8.27  & 7.90  \\
Incorrect        & 3.88  & 16.17 & 8.58  & 7.65  & 10.21 & 22.01 \\
No description   & 6.89  & 0.00  & 0.00  & 0.00  & 0.00  & 0.00  \\
\bottomrule
\end{tabularx}
\caption{Detailed performance metrics of LLMs in generating SQL column descriptions.}
\label{tab:quality_distribution}
\normalsize
\end{table*}

\subsection{Details of Improving the Existing BIRD-Bench Column Descriptions}
\label{app:improve-column-descriptions}

\begin{table*}[h] 
\centering
\small
\begin{tabularx}{0.65\linewidth}{>{\raggedright\arraybackslash}X >{\centering\arraybackslash}p{1.5cm} >{\centering\arraybackslash}p{1.8cm} >{\centering\arraybackslash}p{1.8cm}}
\toprule
\textbf{Label}             & \textbf{(1) BIRD-Bench} & \textbf{(2) LLM Updated} & \textbf{(3) Manually Updated} \\
\midrule
Perfect                    & 40.24           & 70.80              & 100                     \\
Somewhat correct           & 38.80           & 18.80              & 0                       \\
Incorrect                  & 1.88            & 0.06               & 0                       \\
I can't tell               & 0.81            & 7.27               & 0                       \\
No description             & 18.27           & 3.07               & 0                       \\
\midrule
\textbf{Total}             & 100\%             & 100\%                & 100\%                     \\
\bottomrule
\end{tabularx}
\caption{Quality of the original BIRD-Bench column descriptions and the LLM-improved descriptions.}
\label{table:quality_comparison}
\end{table*}

To improve column descriptions, we used GPT-4o, providing it with the BIRD-Bench column descriptions, corresponding schema in \texttt{CREATE\_TABLE} statements, example data rows from the corresponding table, and instructions to generate descriptions for given columns. The model was given clear guidelines and in-context examples of the desired description format. The full prompt template can be found in Appendix \ref{app:prompt}.

Two human annotators, who are authors of this paper, fluent in English and experts in SQL, independently assessed the original BIRD-Bench descriptions and the LLM-generated descriptions. They assigned each description to one of the following categories: \textbf{Perfect}, indicating that the description contains sufficient information for an unambiguous interpretation of the column; \textbf{Somewhat Correct}, where the description is somewhat correct, but there is room for improvement; \textbf{Incorrect}, if the description contains inaccuracies; N\textbf{o Description}, when a description is missing; and \textbf{I can't tell}, when the annotator is unable to determine the accuracy of the description based on the available data.

The full annotator guidelines and more detailed descriptions of the labels are presented in Appendix \ref{app:anno-quality-improved}.

Annotators' assessments of the three datasets are shown in Table \ref{table:quality_comparison}. There was a substantial level of agreement between annotators, with a Cohen's kappa of $0.68$ on the original BIRD-Bench descriptions and $0.61$ on the improved generated descriptions. The GPT-4o descriptions showed a significant improvement over the original BIRD-Bench descriptions. 
While not perfect, LLM-generated descriptions can be a useful tool for improving existing column descriptions when there are already quality issues in the existing descriptions.

\newpage
\subsection{Annotation Procedure and Guidelines}

\subsubsection{Annotating the Quality of the Original BIRD-Bench Dataset Column Descriptions}
\label{app:ann-quality-OG-Bird}

Two annotators independently annotated the column descriptions from the BIRD-Bench dataset, following the predefined annotation guidelines. Before starting the annotation process, an introductory session was held, during which a few samples were annotated collaboratively and discussed with a third party. This session aimed to clarify the decision boundaries within the annotation guidelines. During the annotation process, the annotators had access to the SQLite database and SQLite-viewer, the database name, the table name, the column name, examples of annotations of each category and the metadata files provided by BIRD-Bench. When annotating a column, the annotators were first asked to look at the column name and the table name before reading the column description in the corresponding BIRD-Bench metadata file. After reading the description, the annotators were then asked to also take a look at the corresponding database and table and accompanying data in SQLite Viewer. With this whole picture in mind, the annotators were then instructed to determine the quality of the column description according to the guidelines.

\begin{table}[H]
\centering
\small
\begin{tabularx}{\textwidth}{lX}
\toprule
\textbf{Classification} & \textbf{Description} \\
\midrule
Perfect & A perfect column description should contain enough information so that the interpretation of the column is completely free of ambiguity. It does not need to include any descriptions of the specific values inside the column to be considered perfect. The description should contain information about what table the column is referencing. For example, instead of "The name," we want "The name of the client that made the transaction" if we have a transaction database with columns such as \texttt{NAME}, \texttt{AMOUNT}, and \texttt{DATE} to resolve the ambiguity of what the name refers to. Additionally, the column description should be a full and valid English sentence, with proper grammar, capitalization, and punctuation. For instance, instead of "nationality of drivers" when each instance refers to only one driver, it should be "The nationality of a driver." \\
\midrule
Somewhat Correct & The column description is somewhat correct, but there is room for improvement. \\
\midrule
Incorrect & The column description is incorrect. Contains inaccurate or misleading information. It could still contain correct information, but any incorrect information automatically leads to an incorrect rating. \\
\midrule
No Description & The column description is missing. \\
\midrule
I Can't Tell & It is impossible to tell the class of the description with the given information. \\
\midrule
\textbf{Language} & You are allowed to translate textual data in another language in order to determine the difficulty. The act of translating should not affect the assessment of the difficulty. \\
\midrule
\textbf{Tools} & SQLite Viewer in VSCode and SQLite3 from the command line. Gold column descriptions, database names, table names, and column names from our gold dataset.\\
\bottomrule
\end{tabularx}
\caption{The annotators' guidelines for classifying the original column descriptions provided by the BIRD dataset.}
\label{tab:BIRD_column_desc_annotation_guidelines}
\end{table}

\newpage

\subsubsection{Annotating the Quality of the Improved Column Descriptions}
\label{app:anno-quality-improved}

The same annotation procedure that was used for annotating the quality of the original BIRD-Bench descriptions was used for annotating the improved columns, which can be found in Section \ref{app:ann-quality-OG-Bird}.

\subsubsection{Annotating the Difficulty of Predicting the Column Descriptions for the BIRD Dataset}

This section describes the annotations process used when annotating the difficulty of the columns. During the initial annotation phase, both annotators were given access to the necessary tools, including SQLite Viewer in VSCode and SQLite3 from the command line. Each annotator independently reviewed the database name, table name, column name, example data, gold column description and other columns in the table. Based on this information, they assigned a difficulty level to the task of writing a column description. Annotations were documented in a shared spreadsheet. After the initial annotation phase, the annotators compared their ratings. For columns with matching difficulty ratings, the results were finalized. Discrepancies were discussed, referring to the guidelines and data examples to reach a consensus. If consensus could not be reached, a mediator reviewed the case and made the final decision. The final difficulty ratings were documented in the shared spreadsheet, reflecting the consensus or mediator decisions for all columns creating the final gold standard difficulty annotations.

\begin{table}[H]
\centering
\small
\begin{tabularx}{\textwidth}{lX}
\toprule
\textbf{Difficulty Level} & \textbf{Description} \\
\midrule
Very Hard & Given the database name, the table name, the column name and example data from the database, and other columns in the table, it is impossible to accurately determine what the column description should be. \\
\midrule
Hard & Given the database name, the table name, the column name and example data from the database, and other columns in the table, I am unsure what the column description should be. \\
\midrule
Medium & Given the database name, the table name, the column name and example data from the database, and other columns in the table, I can accurately determine what the column description should be. \\
\midrule
Easy & Given only the table name and the column name, and other columns in the table, I can accurately determine what the column description should be. \\
\midrule
\textbf{Language} & You are allowed to translate textual data in another language in order to determine the difficulty. The act of translating should not affect the assessment of the difficulty. \\
\midrule
\textbf{Tools} & SQLite Viewer in VSCode and SQLite3 from the command line. Gold column descriptions, database names, table names and column names from our gold dataset.\\
\bottomrule
\end{tabularx}
\caption{Annotation Guidelines for Difficulty Ratings.}
\label{tab:difficulty_rankings_guidelines}
\end{table}

\subsubsection{Annotating the Quality of Generated Column Descriptions}
\label{app:anno-quality-pred}

\begin{figure*}[t]
    \centering    
    \includegraphics[scale=0.3]{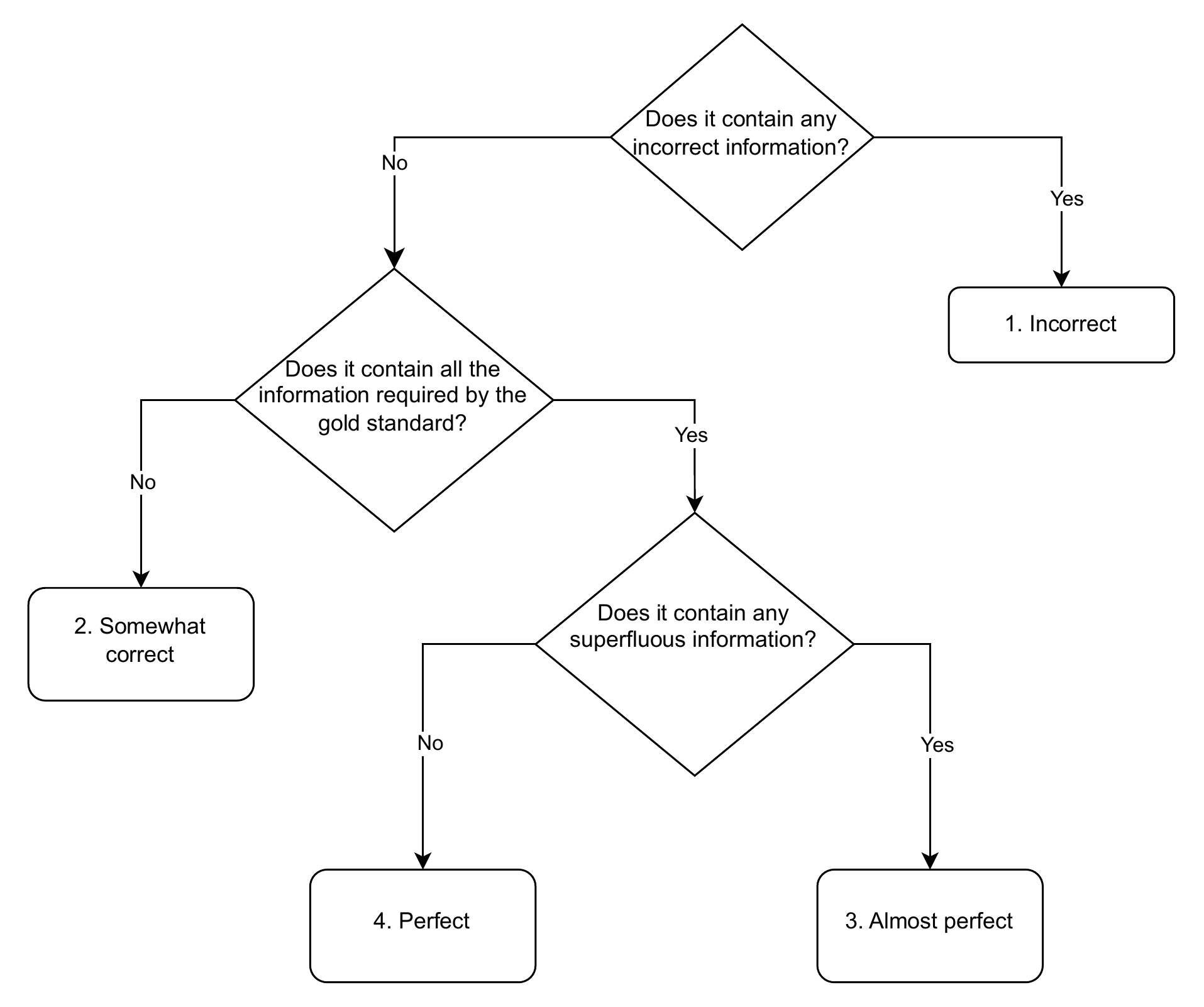} 
    \caption{A decision tree designed to help annotators in deciding the quality of the generated column descriptions.}
    \label{fig:decision}
\end{figure*}

In this section, the annotation procedure to evaluate the quality of the descriptions generated by the different LLMs are described. First, four annotators were organized into two teams of two, with each team assigned to three different LLMs. The guidelines categorized the quality of the column descriptions into five levels: Perfect, Almost Perfect, Somewhat correct, Incorrect, and No description. Before starting the annotation process, an introductory session was held, during which a few samples were annotated collaboratively and discussed with a third party. This session aimed to clarify the decision boundaries within the annotation guidelines. During the annotation process, the annotators had access to the column name, corresponding table and database name, the golden column description, the decision tree depicted in Figure \ref{fig:decision} and the predicted column description. During the annotation process, each annotator independently reviewed the descriptions generated by the LLMs. Annotations were recorded in a shared spreadsheet. Annotation scores were then calculated to quantify the agreement between the annotators and the accuracy of the LLM-generated descriptions relative to the gold standard. 

\begin{table}[H]
\centering
\small
\begin{tabularx}{\textwidth}{lX}
\toprule
\textbf{Quality Level} & \textbf{Description} \\
\midrule
Perfect & Matching the gold description without extra, redundant information. Redundant information is categorized as descriptions that do not provide useful additional information. For example, "<Gold description> + 'is a primary/foreign key'" (NOT REDUNDANT) versus "<Gold description> + 'is useful for retrieving data'" (REDUNDANT). \\
\midrule
Almost Perfect & Matching the gold description but verbose with redundant information, without any incorrect or misleading information. \\
\midrule
Somewhat correct & The column description is somewhat correct but has room for improvement due to missing information. For example, "The Time column records the specific time at which a transaction occurred, formatted in a 24-hour HH:MM pattern," which lacks enough information to make a valid prediction beyond the primary purpose. \\
\midrule
Incorrect & The column description is incorrect and contains inaccurate or misleading information. Any incorrect information automatically leads to an incorrect rating, even if some correct information is present. \\
\midrule
No description & The column description is missing. The models are instructed to output "Not enough information to make a valid prediction" if they cannot generate a suitable column description. \\
\bottomrule
\end{tabularx}
\caption{Annotation Guidelines for LLM Generated Descriptions}
\label{tab:llm_annotation_guidelines}
\end{table}

\newpage
\subsection{Examples of Columns, Gold Descriptions and Difficulties}
\label{app:column_examples}

\begin{table}[H]
\centering
\small
\caption{Illustrative Examples of Column Descriptions by Difficulty}
\label{tab:column_descriptions}
\resizebox{\textwidth}{!}{%
\begin{tabular}{>{\raggedright}p{1.2cm} >{\raggedright}p{2.2cm} >{\raggedright}p{2cm} >{\raggedright}p{2cm} >{\raggedright}p{4cm} >{\raggedright\arraybackslash}p{4cm}}
\toprule
\textbf{Difficulty} & \textbf{Database Name} & \textbf{Table Name} & \textbf{Column Name} & \textbf{Description} & \textbf{Why Ranked This Difficulty} \\
\midrule
Easy & superhero & gender & gender & The name of the gender of the superhero. & Self-explanatory. \\
\midrule
Easy & student\_club & event & event\_date & The date the event took place or is scheduled to take place. & Self-explanatory. \\
\midrule
Medium & student\_club & zip\_code & short\_state & Two-letter state abbreviation. & Requires context: while \texttt{short\_state} suggests state abbreviations, it needs data confirmation to represent two-letter state codes. \\
\midrule
Medium & formula\_1 & results & laps & The number of laps completed by the driver in a race. & Requires basic knowledge of racing and sports. \\
\midrule
Hard & card\_games & foreign\_data & uuid & Unique identifier referencing the related card. & Understanding the concept of a UUID and its relation to the card requires some domain knowledge. \\
\midrule
Hard & formula\_1 & results & rank & The overall rank of the driver's fastest lap time in the race. & Without context, \texttt{rank} could refer to various rankings, making it ambiguous without more info. \\
\midrule
Very Hard & financial & district & A4 & The number of inhabitants in the district. & Column name provides no intuitive clue. Data contains only numbers, offering no indication of its purpose. \\
\midrule
Very Hard & financial & district & A11 & Numeric code for the district. & The column name \texttt{A11} in the \texttt{Districts} table is uninformative; without documentation, it is impossible to infer its meaning. \\
\bottomrule
\end{tabular}%
}
\normalsize
\end{table}

\newpage
\subsection{Prompt Templates}
\label{app:prompt}

\begin{figure}[h]
\begin{lstlisting}[basicstyle=\scriptsize\ttfamily,breaklines=true,tabsize=2,frame=single,escapechar=|]
"""
### TASK:
Context - Generate Column Description for Database, to give users an easier 
time understanding what data is present in the column.

Database Schema Details:
"{database_schema}"

Here is example data from the table {table}: "{example_data}"

Here is up to 10 possible unique values for the column {column} from the 
table {table}: {unique_data}

The column name for {column} is {column_name}. This is the name of the column,
it can contain important information about the column, and should be used to 
write the description. The previous column description is {column_description}. 
This is the old description of the column, this is sometimes lacking and should 
be read and rewritten.

### Task
Generate a precise description for the {column} column in the {table} table. 
Your description should include:
- Primary purpose of the column. If the details in the schema do not suffice 
  to ascertain what the data is, return: "Not enough information to make a valid 
  prediction."
Optionally, your description could also include:
- Additional useful information (if apparent from the schema), formatted 
  as a new sentence, but never more than one. If no useful information is 
  available, return nothing.

### Requirements
- Focus solely on confirmed details from the provided schema.
- Keep the description concise and factual.
- Exclude any speculative or additional commentary.
- DO NOT return the phrase "in the {table} table" in your description.

**Examples:**
- "no. of municipalities with inhabitants < 499": 
  "This is the number of municipalities with fewer than 499 inhabitants."
- Short: "Details about the ratio of urban inhabitants."
- "Frequency": "The frequency of transactions on the account."
- "amount of money": "The amount of money in the order."

### Skip the "data_format" and focus solely on updating the 
"column_description".

DO NOT return anything else except the generated column description.
"""
\end{lstlisting}
\end{figure}

\begin{figure}[H]
\begin{lstlisting}[basicstyle=\scriptsize\ttfamily,breaklines=true,tabsize=2,frame=single,escapechar=|]
"""
### Context - Generate Column Description for Database, to give users an easier 
time understanding what data is present in the column.

Database Schema Details:
""
{database_schema}
""

Here is example data from the table {table}: ""

{example_data}

Here is up to 10 possible unique values for the column {column} from the table 
{table}:

{unique_data}

""

### Task
Generate a precise description for the {column} column in the {table} table. 
Your description should include:
- Primary purpose of the column. If the details in the schema do not suffice 
to ascertain what the data is, return: "Not enough information to make a 
valid prediction."
Optionally, your description could also include:
- Additional useful information (if apparent from the schema), formatted as 
a new sentence, but never more than one. If no useful information is 
available or if the details in the schema do not suffice to ascertain useful 
details, return nothing.

### Requirements
- Focus solely on confirmed details from the provided schema.
- Keep the description concise and factual.
- Exclude any speculative or additional commentary.
- DO NOT return the phrase "in the {table} table" in your description. 
This very important.

DO NOT return anything else except the generated column description. 
This is very important. The answer should be only the generated text aimed at
describing the column.
"""
\end{lstlisting}
\caption{The prompt used for generating columns, given only the schema and sample data from the database.} 
\label{col_gen_template}
\end{figure}

\begin{figure}[H]
\begin{lstlisting}[basicstyle=\scriptsize\ttfamily,breaklines=true,tabsize=2,frame=single,escapechar=|]
"""
### TASK:
Context - You are an expert in converting natural language questions and 
instructions into their corresponding SQL counterpart.

Database schema and the associated column descriptions for each table:

{database_schema}

Using valid SQL, answer the following question based on the tables provided 
above by converting the natural language question into the corresponding SQL
query.

Question: {question}

### Requirements
DO NOT return anything else except the SQL query. Do not think out loud. 
ONLY return the SQL query, nothing else.
"""
\end{lstlisting}
\caption{Zero-Shot Prompting Template.} 
\label{zero_shot_template}
\end{figure}

\begin{figure}[H]
\begin{lstlisting}[basicstyle=\scriptsize\ttfamily,breaklines=true,tabsize=2,frame=single,escapechar=|]
"""
You are evaluating a response that has been submitted for a particular task, 
using a specific set of standards. Below is the data:  
[BEGIN DATA] 
*** 
[Task]: The task is to generate accurate descriptions of columns in SQL 
databases, given only access to the schema in a CREATE_TABLE format and example 
rows from the database.  

The goal is to create informative descriptions which reduces ambiguity and 
increases understanding for users of the database.  

*** 
[Submission]: {response} 
*** 
[Gold Answer]:  {gold_answer} 
*** 
[Criterion]: Evaluation Criteria

Correctness:  
4: Perfect (Matching the GOLD description or better): 

Matching the gold description without extra, redundant 
information. To redundant information, descriptions which do not provide useful 
additional information, is categorized.  

Example: <Gold description> + "is a primary/foreign key." can be considered 
useful so the extra information is NOT REDUNDANT.  

Gold description" + "is useful for retrieveing data" does not any extra useful 
information so is considered REDUNDANT. 

3: Almost Perfect 

Matching the GOLD description but is verbose with extra, redundant information 
(but all the information is correct, ie it does not contain any incorrect, 
misleading information). 

2: Somewhat correct 

The column description is somewhat correct, but there is room for improvement 
due to missing information.  

Example: "The Time column records the specific time at which a transaction 
occurred, formatted in a 24-hour HH:MM:SS pattern. Not enough information to 
make a valid prediction beyond the primary purpose." 

1: Incorrect 

The column description is incorrect. Contains innacurrate or misleading 
information. Could still contain correct information but any incorrect 
information automatically leads to an incorrect rating.  

*** 
[END DATA] 

Does the submission meet the criterion? First, write out in a step by step 
manner your reasoning about the criterion to be sure that your conclusion is 
correct. Avoid simply stating the correct answers at the outset.  

Your response must be RFC8259 compliant JSON following this schema: 

{{"reasoning": str, "correctness": int}} 

Make sure your output is a valid json string in the format provided above. 
"""
\end{lstlisting}
\caption{Template for the LLM-as-a-judge experiments.} 
\label{judge_template}
\end{figure}

\end{document}